\documentclass[10pt,twocolumn,letterpaper]{article}

\usepackage{cvpr}              % To produce the CAMERA-READY version
\usepackage{graphicx}
\usepackage{amsmath}
\usepackage{amssymb}
\usepackage{booktabs}
\usepackage{float}
\usepackage{color}
\usepackage{multirow}
\usepackage{subcaption}
\usepackage{color}
\usepackage{xcolor}
\usepackage{pifont}
\usepackage[accsupp]{axessibility}

\definecolor{bus_color}{RGB}{225, 80, 23}
\definecolor{truck_color}{RGB}{200, 110, 95}
\definecolor{pedestrian_color}{RGB}{37, 31, 105}
\definecolor{car_color}{RGB}{236, 152, 14}

\usepackage[pagebackref,breaklinks,colorlinks]{hyperref}

\newcommand{\mycirc}[1][black]{\Large\textcolor{#1}{\ensuremath\bullet}}

\usepackage[capitalize]{cleveref}
\crefname{section}{Sec.}{Secs.}
\Crefname{section}{Section}{Sections}
\Crefname{table}{Table}{Tables}
\crefname{table}{Tab.}{Tabs.}

\begin{document}

\title{Cross Modal Transformer: Towards Fast and Robust 3D Object Detection}

\author{Junjie Yan \qquad Yingfei Liu \qquad Jianjian Sun \qquad Fan Jia \qquad Shuailin Li \\ \qquad Tiancai Wang \qquad Xiangyu Zhang \vspace{1mm}\\ MEGVII Technology\\
}

\maketitle

%%%%%%%%% ABSTRACT
\begin{abstract}
   In this paper, we propose a robust 3D detector, named Cross Modal Transformer (CMT), for end-to-end 3D multi-modal detection. Without explicit view transformation, CMT takes the image and point clouds tokens as inputs and directly outputs accurate 3D bounding boxes. The spatial alignment of multi-modal tokens is performed by encoding the 3D points into multi-modal features. The core design of CMT is quite simple while its performance is impressive. It achieves 74.1\% NDS (state-of-the-art with single model) on nuScenes test set while maintaining fast inference speed. Moreover, CMT has a strong robustness even if the LiDAR is missing. Code is released at \url{https://github.com/junjie18/CMT}.
\end{abstract}

% \let\thefootnote\relax\footnotetext{{\large\ding{41}} Corresponding author.}

%%%%%%%%% BODY TEXT
\section{Introduction}
Multi-sensor fusion has shown its great superiority in autonomous driving system~\cite{liu2022bevfusion,chen2022futr3d, li2022unifying, bai2022transfusion, liang2022bevfusion}. Different sensors usually provide the complementary information for each other. For instance, the camera captures information in a perspective view and the image contains rich semantic features while point clouds provide much more localization and geometry information. Taking full advantage of different sensors helps reduce the uncertainty and makes accurate and robust prediction. 

\begin{figure}[t]
	\centering  
\includegraphics[width=0.475\textwidth]{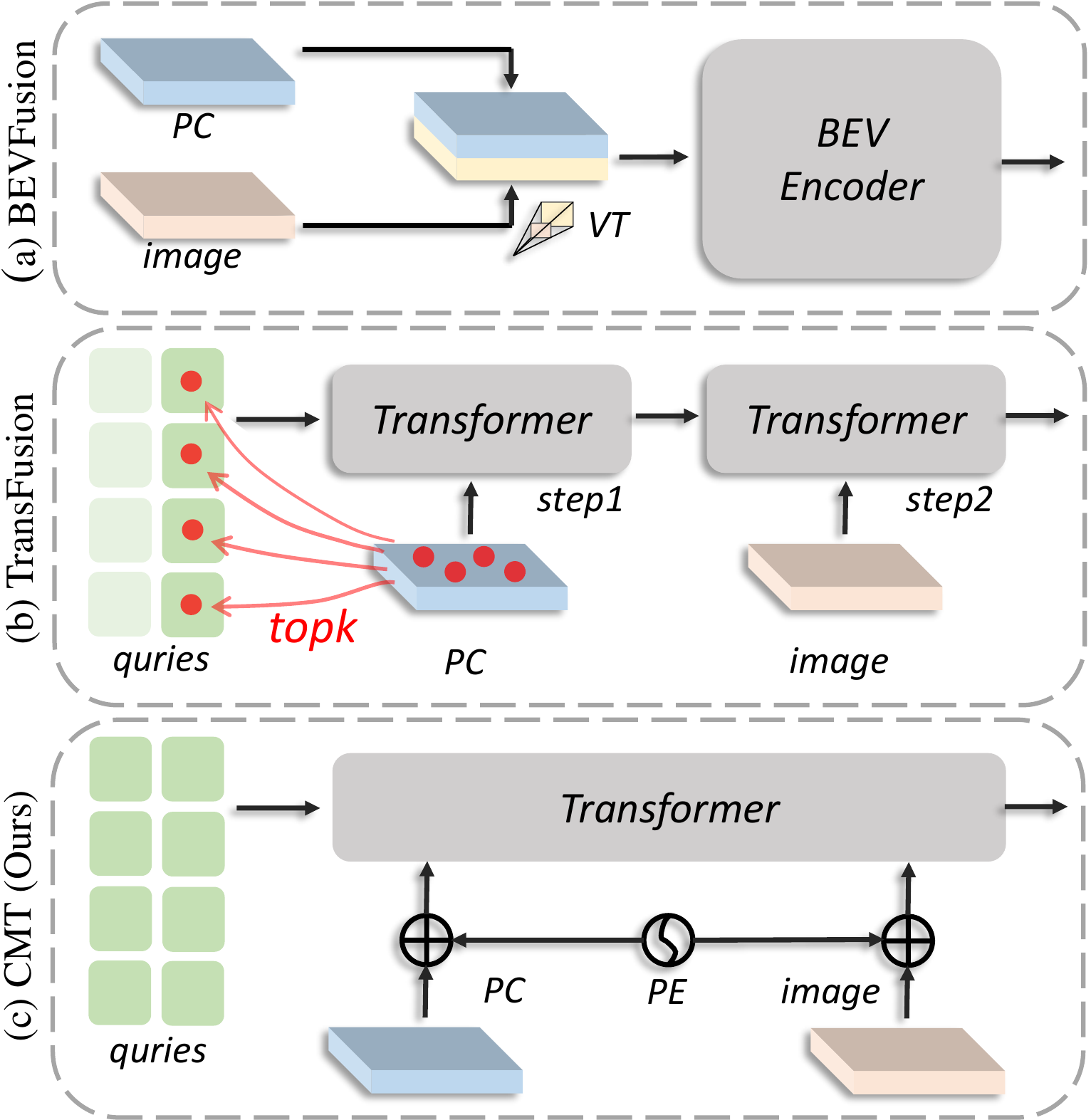}
	\caption{Comparison between BEVFusion, TransFusion, and our proposed CMT. (a) In BEVFusion, 
 the camera features are transformed into BEV space by view transform. Two modality features are concatenated in BEV space and the BEV encoder is adopted for fusion. (b) TransFusion first generates the queries from the high response regions of LiDAR features. After that, object queries interact with point cloud features and image features separately. (c) In CMT, the object queries directly interact with multi modality features simultaneously. Position encoding (PE) is added to the multi-modal features for alignment. "VT" is the view transformation from image to 3D space.
   }  
\label{fig:intro}
\end{figure}

Sensor data of different modalities usually has large discrepancy in distribution, making it hard to merge the multi-modalities. State-of-the-art (SoTA) methods tend to fuse the multi-modality by constructing unified bird’s-eye-view (BEV) representation~\cite{liu2022bevfusion,liang2022bevfusion,li2022unifying} or querying from tokens~\cite{bai2022transfusion,chen2022futr3d}.
For example, BEVFusion~\cite{liu2022bevfusion} explores a unified representation by BEV transformation for BEV feature fusion (see \cref{fig:intro}(a)). TransFusion~\cite{bai2022transfusion} follows a two-stage pipeline and the camera images in second stage provide supplementary information for prediction refinement (see \cref{fig:intro}(b)). However, exploring a truly end-to-end pipeline for multi-sensor fusion remains to be a question.

\label{sec:intro}

\begin{figure*}[t]
\centering

\begin{tabular}{cc}
% \hspace{-0.7em}
\includegraphics[width=0.465\textwidth]{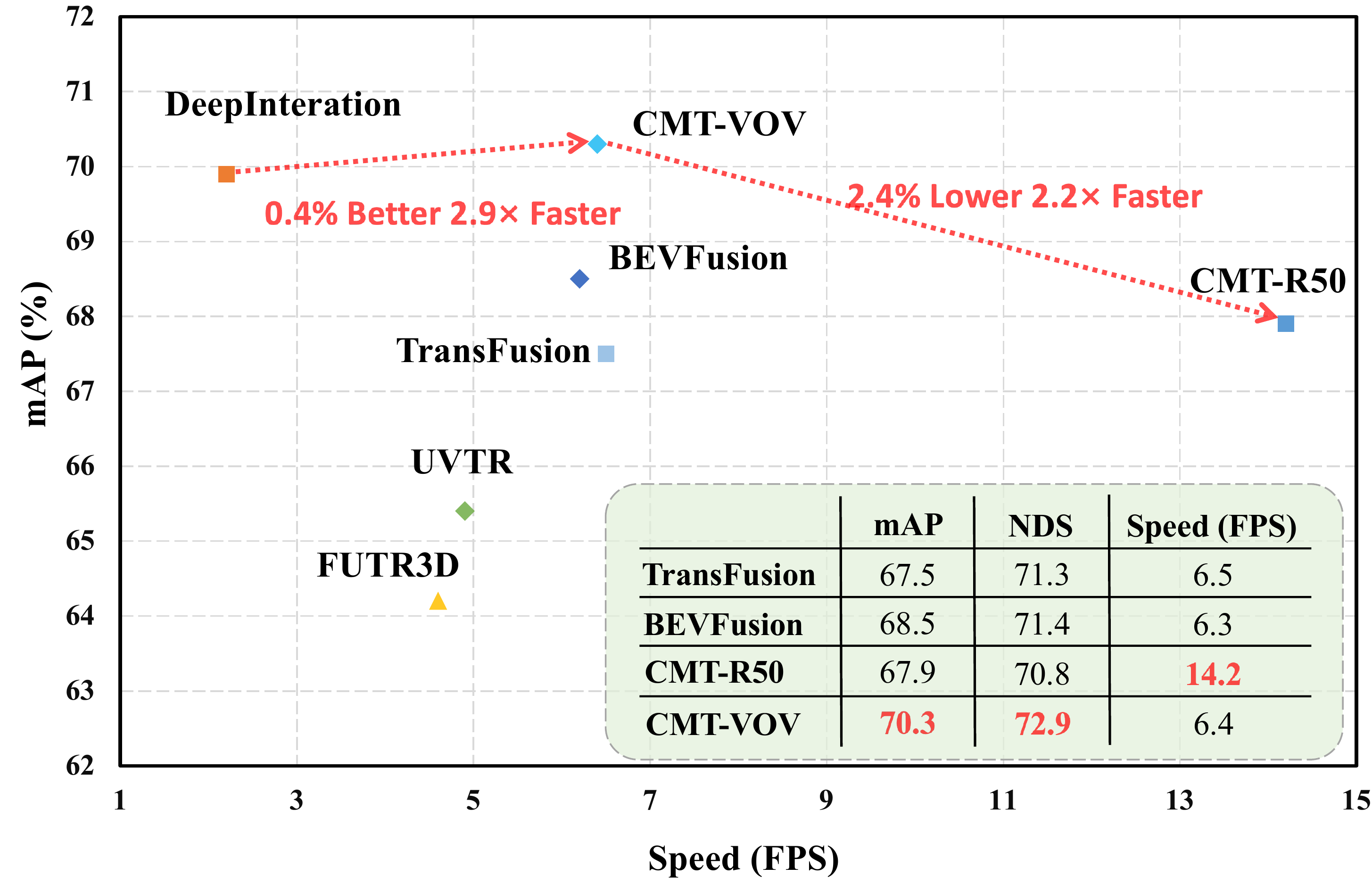} &
\hspace{-0.5em} 
\includegraphics[width=0.445\textwidth]{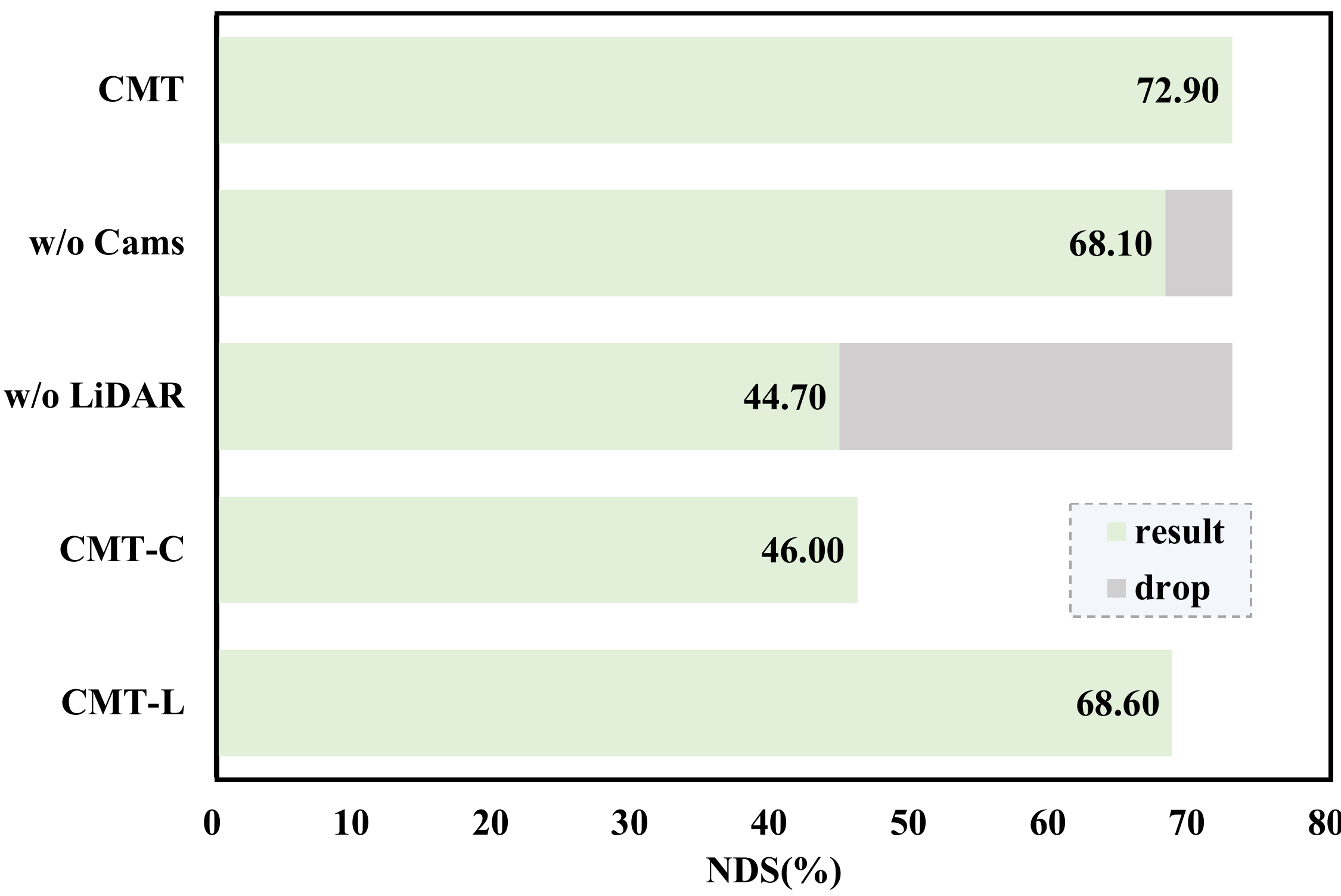}\\
% (a) FPS and Performace. & (b) Evaluation on sensor missing. \\
\end{tabular}
\vspace{-0.3cm}
\caption{\textbf{Left}: Performance comparison between CMT and existing methods. 
% CMT achieves state-of-the-art performance with much higher inference speed.
All speed statistics are measured on a single Tesla A100 GPU using the best model of official repositories. (All methods use the same Voxelization module from \href{https://github.com/traveller59/spconv}{spconv} repo. Moreover,  TranFusion head of BevFusion is also equipped with the same \href{https://github.com/Dao-AILab/flash-attention}{FlashAttn} in \href{https://github.com/junjie18/CMT}{CMT} repo. Both CMT and BevFusion haven't adopted precompute~\cite{liu2022bevfusion}).
\textbf{Right}: Performance evaluation of CMT under sensor missing. During inference, CMT achieves vision-based performance when LiDAR is missing, showing strong robustness.}
\label{fig:fps_robust}
\vspace{-.1in}
\end{figure*}

Recently, the effectiveness of end-to-end object detection with transformer (DETR)~\cite{carion2020detr,zhu2020deformable} has been proved in many perception tasks, such as instance segmentation~\cite{dong2021solq,queryinst2021}, multi-object tracking~\cite{zeng2022motr,Meinhardt2021trackformer} and visual 3D detection~\cite{wang2022detr3d,liu2022petr,liu2022petrv2}. The DETR architecture is simple yet effective thanks to the object queries for representing different instances and bipartite matching for one-to-one assignment. 

Inspired by DETR, we aim to build an elegant end-to-end pipeline for multi-modal fusion in 3D object detection.
In DETR, object queries directly interact with the image tokens through cross-attention in transformer decoder. For 3D object detection, one intuitive way is to concatenate the image and point cloud tokens together for further interaction with object queries. However, the concatenated tokens are disordered and unaware of their corresponding locations in 3D space. Therefore, it is necessary to provide the location prior for multi-modal tokens and object queries.

% \begin{figure}[t]
% 	\centering  
%     \includegraphics[width=0.48\textwidth]{cmt_fps.pdf}
%     \vspace{-0.5cm}
% 	\caption{CMT obtains SoTA performance among all existing methods, while simultaneously maintain the fastest speed. All speed statistics are evaluated on a single Tesla A100 GPU. We report the best public performance of each model on the nuScenes \textbf{val} set in this figure.}
% \label{fig:speed_result}
% \end{figure}

In this paper, we propose Cross-Modal Transformer (CMT), a simple yet effective end-to-end pipeline for robust 3D object detection (see \cref{fig:intro}(c)). 
First, we propose the Coordinates Encoding Module (CEM), which produces position-aware features, by encoding 3D points set implicitly into multi-modal tokens. Specifically, for camera images, 3D points sampled from frustum space are used to indicate the probability of 3D positions for each pixel. While for LiDAR, the BEV coordinates are simply encoded into the point cloud tokens.
Next, we introduce the position-guided queries. Each query is initialized as a 3D reference point following PETR~\cite{liu2022petr}. We transform the 3D coordinates of reference points to both image and LiDAR spaces, to perform the relative coordinates encoding in each space. 
% Moreover, for faster convergence, we introduce the inductive bias of locality, by extending Query Denoising~\cite{li2022dn} to a point-based formulation. 

The proposed CMT framework brings many advantages compared to existing methods. Firstly, our method is a simple and end-to-end pipeline and can be easily extended. The 3D positions are encoded into the multi-modal features implicitly, which avoids introducing the bias caused by explicit cross-view feature alignment. Secondly, our method only contains basic operations, without the feature sampling or complex 2D-to-3D view transformation on multi-modal features. It achieves state-of-the-art performance and shows obvious superiority compared to existing approaches, as shown in the left graph of \cref{fig:fps_robust}. Thirdly, the robustness of our CMT is much stronger than other existing approaches. Extremely, under the condition of LiDAR miss, our CMT with only image tokens can achieve similar performance compared to those vision-based 3D object detectors~\cite{liu2022petr,li2022bevformer} (see the right graph of \cref{fig:fps_robust}).

% \begin{figure}[t]
% 	\centering  
%     \includegraphics[width=0.48\textwidth]{robust.pdf}
%     \vspace{-0.5cm}
% 	\caption{CMT has a strong robustness under sensor missing condition. During inference, CMT without LiDAR achieves similar detection performance compared to the SoTA camera-only detector PETR\cite{liu2022petr}. CMT without camera input only introduce a slight drop, compared to our LiDAR-only baseline CMT-L. (Note: we evaluate without any finetune process)}
% \label{fig:robust_result}
% \end{figure}

To summarize, our contributions are: 
\begin{itemize}
\item we propose a fast and robust 3D detector, which is a truly end-to-end framework without any post-process. It overcomes the sensor missing problem.
\item The 3D positions are encoded into the multi-modal tokens, without any complex operations, like grid sampling and voxel-pooling. 
\item CMT achieves state-of-the-art 3D detection performance on nuScenes dataset. It provides a simple baseline for future research.
\end{itemize}

%------------------------------------------------------------------------
\section{Related Work}

\subsection{Camera Based 3D Object Detection}
Camera-based 3D object detection is one of the basic tasks in computer vision. Early works~\cite{wang2021fcos3d,wang2022pgd} mainly follow the dense prediction pipeline. They first localize the objects on image plane and then predict their relevant 3D attributes, such as depth, size and orientation. However, with the surrounding cameras, the perspective-view based design requires elaborate post-processes to eliminate the redundant predictions of the overlapping regions. 
Recently, 3D object detection under the BEV has attracted increasing attention. The BEV representation provides a unified coordinate to fuse information from multiple camera views. LSS~\cite{philion2020lift}, BEVDet ~\cite{huang2021bevdet} and BEVDepth~\cite{li2022bevdepth} predict the depth distribution to lift the image features to 3D frustum meshgrid. Besides, inspired by DETR~\cite{carion2020end}, DETR3D~\cite{wang2022detr3d} and BEVFormer~\cite{li2022bevformer} project the predefined BEV queries onto images and then employ the transformer attention to model the relation of multi-view features. The above methods explicitly project the local image feature from 2D perspective view to BEV. Different from them, PETR~\cite{liu2022petr,liu2022petrv2} and SpatialDETR~\cite{doll2022spatialdetr} adopt the positional embedding that depends on the camera poses, allowing the transformer to implicitly learn the projection from image views to 3D space.

\begin{figure*}[t]
	\centering  
\includegraphics[height=5.0cm,width=17.5cm]{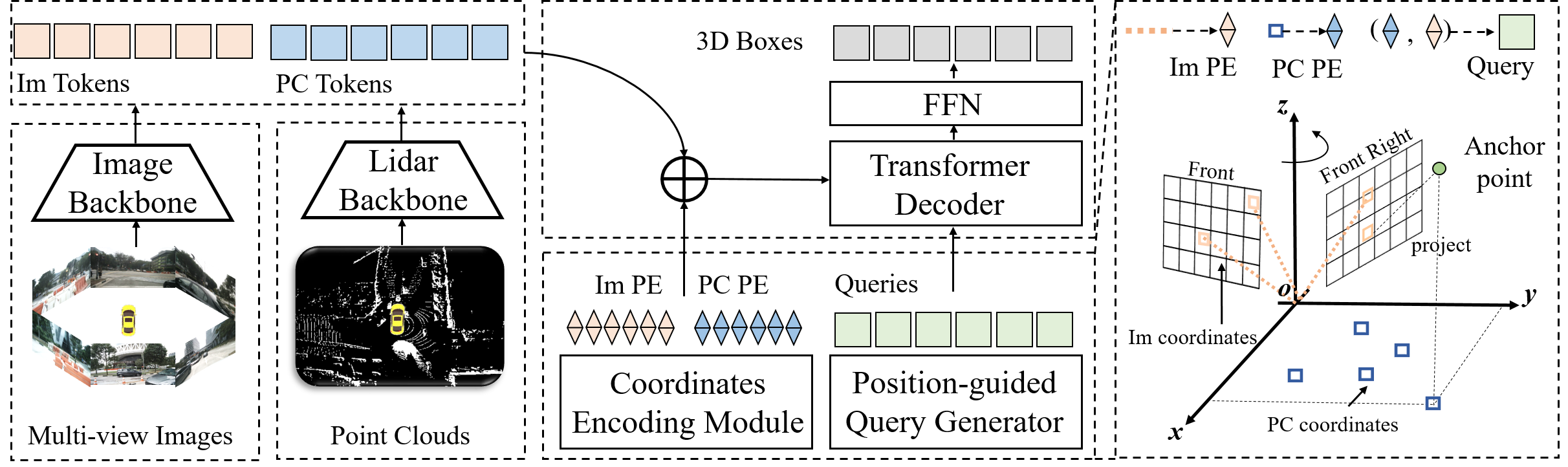}\vspace{-0.2cm}
\caption{The architecture of Cross-Modal Transformer (CMT) paradigm. The multi-view images and point clouds are input to two backbone networks to extract feature tokens. In coordinates encoding module, coordinates of camera rays and BEV positions are transformed into the image position encoding (Im PE) and point cloud position encoding (PC PE), respectively. 
The queries are generated by the position-guided query generator. 
In query generator, 3D anchor points are projected to different modalities and the relative coordinates are encoded (see the right part). 
Multi-modal tokens further interact with queries in the transformer decoder. The updated queries are further used to predict the 3D bounding boxes. 
% To accelerate the model convergence, the point-based query denoising is introduced.  
}  
\label{fig:architecture}
\end{figure*}

\subsection{LiDAR Based 3D Object Detection}
LiDAR-based 3D object detection aims to predict 3D object bounding boxes using the point clouds captured from LiDAR. Existing methods process the point cloud into different representations.
Point-based methods~\cite{qi2018frustum,qi2017pointnet,qi2017pointnet++,shi2019pointrcnn,li2021lidar,yang20203dssd} directly extract features from raw point clouds and predict 3D bounding boxes.
PointNet~\cite{qi2017pointnet} is the first architecture to process the point cloud in an end-to-end manner, which preserves the spatial characteristics of the point cloud.
Other methods project the unordered, irregular LiDAR point clouds onto a regular feature space such as 3D voxels~\cite{zhou2018voxelnet, yan2018second, chen2022focal, chen2023largekernel3d}, feature pillars~\cite{lang2019pointpillars,wang2020pillar,yin2021center} and range images~\cite{fan2021rangedet,sun2021rsn}. Then the features are extracted in the BEV plane using a standard 2D backbone.  VoxelNet~\cite{zhou2018voxelnet} first divides the raw point clouds into regular voxel grids, and then uses PointNet network to extract features from the points in each voxel grid.

\subsection{Multi-modal 3D Object Detection}
Multi-sensor fusion in 3D detection has gained great attention in recent years.
State-of-the-art (SoTA) methods tend to find a unified representation for both modalities, or define object queries to fuse the features for further prediction. For example, BEVFusion\cite{liu2022bevfusion, liang2022bevfusion} applies a lift-splat-shoot (LSS) operation to project image feature onto BEV space and concatenates it with LiDAR feature. UVTR\cite{li2022unifying} generates a unified representation in the 3D voxel space by deformable attention\cite{zhu2020deformable}. While for query-based methods, FUTR3D\cite{chen2022futr3d} defines the 3D reference points as queries and directly samples the features from the coordinates of projected planes. TransFusion\cite{bai2022transfusion} follows a two-stage pipeline. The proposals are generated by LiDAR features and further refined by querying the image features. 

\subsection{Transformer-based Object Detection}
The pioneering work DETR~\cite{carion2020detr} proposes a transformer-based detector paradigm without any hand-craft components, and has achieved state-of-the-arts in both 2D and 3D detection~\cite{zhang2022dino, Chen2022GroupDV, li2022bevformer, liu2022petrv2}. However, DETR-like methods usually suffer from  the slow convergence. To this end, many works~\cite{zhu2020deformable, Zhang_2022_CVPR, liu2022dab, li2022dn, zhang2022dino, chen2022group, jia2022detrs} are proposed to improve the training efficiency from various aspects.
Other improvements in 2D detection mainly focus on modifying the transformer layers\cite{zhu2020deformable, Zhang_2022_CVPR}, designing informative object queries\cite{liu2022dab, li2022dn, zhang2022dino}, or exploring the label assignment mechanism\cite{chen2022group, jia2022detrs}.
Deformable DETR\cite{zhu2020deformable} proposes the deformable attention, which only attends to sampling points of local regions.
SAM-DETR\cite{Zhang_2022_CVPR} presents a semantic aligner between object queries and encoded features to accelerate the matching process.
To alleviate the instability of bipartite matching, DAB-DETR\cite{liu2022dab} formulates the object queries as dynamic anchor boxes, while DN-DETR\cite{li2022dn} auxillarily reconstructs the ground-truths from the noisy ones. Based on them, DINO\cite{zhang2022dino} further improves the denoising anchor boxes via a contrastive way.

%-------------------------------------------------------------------------
\section{Method} 
The overall architecture of the proposed CMT is illustrated in \cref{fig:architecture}.  Multi-view images and LiDAR points are fed into two individual backbones to extract multi-modal tokens. 
The 3D coordinates are encoded into the multi-modal tokens by the \emph{coordinates encoding}. The queries from the \emph{position-guided query generator} are used to interact with the multi-modal tokens in transformer decoder and then predict the object class as well as the 3D bounding boxes. 
% \emph{Point-based query denoising} is further introduced to accelerate the training convergence by introducing local prior.
The whole framework is learned in a fully end-to-end manner and the LiDAR backbone is trained from scratch without pretraining.

\subsection{Coordinates Encoding Module}
The coordinates encoding module (CEM) is used to encode the 3D position information into multi-modal tokens. It generates both the camera and BEV position encodings (PEs), which are added to image tokens and point cloud tokens respectively. With the help of CEM, multi-modal tokens can be implicitly aligned in 3D space. 

Let $P(u, v)$ be the 3D points set corresponding to the feature map $F(u, v)$ of different modalities. Here $(u, v)$ indicates the coordinate in the feature map. Specifically, $F$ is the image feature for camera while BEV feature for LiDAR. Suppose the output position embedding of CEM is $\Gamma(u, v)$, its calculation can be formulated as:
\begin{equation}
\begin{aligned}
% \mathit{PE(u, v)} = \psi(P(u, v))
\mathit{\Gamma(u, v)} = \psi(P(u, v))
\end{aligned}
\label{eq:ce}
\end{equation}
where $\psi$ is a multi-layer perception (MLP) layer.
% Since $P(u, v)$ is a set of points, we flatten the coordinates here. 

\noindent \textbf{CE for Images.} Since the image is captured from a perspective view, each pixel can be seen as an epipolar line in 3D space. Inspired by PETR~\cite{liu2022petr}, for each image, we encode a set of points in camera frustum space to perform the coordinates encoding. Given the image feature $F_{im}$,
% suppose $(u, v), 1\leq u \leq H_F, 1\leq v \leq W_F$ is the pixel coordinate in the image feature map, 
each pixel can be formulated as a series of points $\{p_k(u, v) = (u * d_k, v * d_k, d_k, 1)^T, k=1,2,...,d\}$ in the camera frustum coordinates. Here, $d$ is the number of points sampled along the depth axis. The corresponding 3D points can be calculated by:
\begin{equation}
p_k^{im}(u, v) = T_{c_i}^l K_i^{-1} p_k(u, v)
\label{eq:camera2lidar}
\end{equation}
where $T_{c_i}^l \in R^{4\times 4}$ is the transformation matrix from the $i$-th camera coordinate to the LiDAR coordinate. $K_i \in 4\times 4$ is the intrinsic matrix of $i$-th camera. The position encoding of pixel $(u, v)$ for image is formulated as:
\begin{equation}
\begin{aligned}
\mathit{\Gamma_{im}}(u, v) = \psi_{im}(\{p_k^{im}(u, v), \quad k=1,2,...,d\})
\end{aligned}
\label{eq:cameraCoordinatesEncoding}   
\end{equation}
\noindent \textbf{CE for Point Clouds.} We choose VoxelNet\cite{yan2018second, zhou2018voxelnet} or PointPillar\cite{lang2019pointpillars} as backbone to encode the point cloud tokens $F_{pc}$. Intuitively, the point set $P$ in \cref{eq:ce} can be sampled along the Z-axis. Suppose $(u, v)$ is the coordinates in BEV feature map, the sampled point set is then $p_k(u, v)=(u, v, h_k, 1)^T$, where $h_k$ indicates the height of $k$-th points and $h_0=0$ as default. The corresponding 3D points of BEV feature map can be calculated by:
\begin{equation}
\begin{aligned}
p_k^{pc}(u, v) =& (u * u_d, v * v_d, h_k, 1) \\
\end{aligned}
\end{equation}
where $(u_d, v_d)$ is the size of each BEV feature grid. To simplify, we only sample one point along the height axis. It is equivalent to the 2D coordinate encoding in BEV space. The position embedding of point cloud can be obtained by:
\begin{equation}
\begin{aligned}
\mathit{\Gamma_{pc}}(u, v) = \psi_{pc}(\{p_k^{pc}(u, v), \quad k=1,2,...,h\})
\end{aligned}
\label{eq:lidarCoordinatesEncoding}   
\end{equation}

\subsection{Position-guided Query Generator}
Following Anchor-DETR~\cite{wang2021anchor} and PETR~\cite{liu2022petr}, we firstly initialize the queries with $n$ anchor points $A = \{a_i = (a_{x,i}, a_{y,i}, a_{z,i}), i=1,2,...,n\}$ sampled from uniform distribution between $[0, 1]$. Then these anchor points are transformed into 3D world space by linear transformation:
\begin{equation}\label{eq2}
\left\{
\begin{aligned}
&a_{x,i} = &a_{x,i} * &(x_{max}-x_{min}) + x_{min}\\
&a_{y,i} = &a_{y,i} * &(y_{max}-y_{min}) + y_{min}\\
&a_{z,i} = &a_{z,i} * &(z_{max}-z_{min}) + z_{min}
\end{aligned}
\right.
\end{equation}
where $[x_{min},y_{min},z_{min},x_{max},y_{max},z_{max}]$ is the region of interest (RoI) of 3D world space. After that, we project the 3D anchor points $A$ to different modalities and encode the corresponding point sets by CEM.
Then the positional embedding $\Gamma_{q}$ of object queries can be generated by:
\begin{equation}
    \mathit{\Gamma_{q}} = \psi_{pc}(A_{pc}) + \psi_{im}(A_{im})
\end{equation}
where $A_{pc}$ and $A_{im}$ are the point set projected on BEV plane and image plane, respectively.
The positional embedding $\Gamma_{q}$ are further added with the query content embedding to generate the initial position-guided queries $Q_{0}$.

\begin{table*}[t]
\begin{center}
\caption{Performance comparison on the nuScenes \textbf{test} set. ``L" is LiDAR and ``C" is camera.}
\vspace{-0.2cm}
\label{tab:nuscenes_test}
\setlength{\tabcolsep}{2pt}
\begin{tabular}{l|c|cc|ccccc}
\hline\noalign{\smallskip}
Methods & Modality & NDS$\uparrow$ & mAP$\uparrow$ & mATE$\downarrow$ & mASE$\downarrow$ & mAOE$\downarrow$ & mAVE$\downarrow$ & mAAE$\downarrow$ \\

\noalign{\smallskip}
\hline
\noalign{\smallskip}
BEVDet~\cite{huang2021bevdet} & C &0.488 &0.424 &0.524 &0.242 &0.373 &0.950 &0.148  \\
DETR3D~\cite{wang2022detr3d} & C &0.479 &0.412  &0.641 &0.255 &0.394 &0.845 &0.133  \\
PETR~\cite{liu2022petr} & C &0.504 &0.441 &0.593 &0.249 &0.383 &0.808 &0.132 \\
\midrule

CenterPoint~\cite{yin2021center} & L & 0.673 & 0.603 & 0.262 & 0.239 & 0.361 & 0.288 & 0.136 \\
UVTR~\cite{li2022unifying} & L & 0.697 & 0.639 & 0.302 & 0.246 & 0.350 & 0.207 & 0.123 \\ 
TransFusion~\cite{bai2022transfusion} & L & 0.702 & 0.655 & 0.256 & 0.240 & 0.351 & 0.278 & 0.129 \\
\midrule

PointPainting\cite{vora2020pointpainting}~ & LC & 0.610& 0.541 & 0.380 & 0.260 & 0.541 & 0.293 & 0.131\\
PointAugmenting\cite{wang2021pointaugmenting}~ & LC & 0.711 & 0.668 & 0.253 & 0.235 & 0.354 & 0.266 & 0.123\\
MVP\cite{chen2017multi}~ & LC & 0.705 & 0.664& 0.263& 0.238 & 0.321 & 0.313 & 0.134\\
FusionPainting\cite{xu2021fusionpainting}~ & LC & 0.716 & 0.681 & 0.256 & 0.236 & 0.346 & 0.274 & 0.132\\
UVTR~\cite{li2022unifying} & LC & 0.711 & 0.671 & 0.306 & 0.245 & 0.351 & 0.225 & 0.124\\
TransFusion~\cite{bai2022transfusion} & LC & 0.717 & 0.689 & 0.259 & 0.243 & 0.359 & 0.288 & 0.127 \\
BEVFusion~\cite{liu2022bevfusion} & LC & 0.729 & 0.702 & 0.261 & 0.239 & 0.329 & 0.260 & 0.134 \\
DeepInteration~\cite{yang2022deepinteraction} & LC & 0.734 & 0.708 & 0.257 & 0.240 & 0.325 & 0.245 & 0.128 \\
\midrule
CMT-C~ & C & 0.481 & 0.429 & 0.616 & 0.248 & 0.415 &0.904 & 0.147 \\
CMT-L~ & L & \textbf{0.701} & \textbf{0.653} & 0.286 & 0.243 & 0.356 & 0.238 & 0.125 \\
CMT~ & LC & \textbf{0.741} & \textbf{0.720} & 0.279 & 0.235 & 0.308 & 0.259 & 0.112 \\
\bottomrule
\end{tabular}
\end{center}
% \vspace{-1.5em}
\end{table*}

\begin{table}[t]
\begin{center}
\caption{Performance comparison on the nuScenes \textbf{val} set. ``L" is LiDAR and ``C" is camera.}\vspace{-0.2cm}
\label{tab:nuscenes_val}
\begin{tabular}{l|c|cc}
\hline\noalign{\smallskip}
Methods & modality & NDS$\uparrow$ & mAP$\uparrow$ \\
\midrule
FUTR3D~\cite{chen2022futr3d} & L & 0.655 & 0.593 \\
UVTR~\cite{li2022unifying} & L & 0.676 & 0.608 \\
TransFusion~\cite{bai2022transfusion} & L & 0.701 & 0.651 \\
\midrule
FUTR3D~\cite{chen2022futr3d} & LC &0.683 &0.645 \\
UVTR~\cite{li2022unifying} & LC &0.702 &0.654 \\
TransFusion~\cite{bai2022transfusion} & LC & 0.713 & 0.675 \\
BEVFusion~\cite{liu2022bevfusion} & LC & 0.714 & 0.685 \\
DeepInteration~\cite{yang2022deepinteraction} & LC & 0.726 & 0.699 \\
\midrule
CMT-C & C & 0.460 & 0.406 \\
CMT-L & L & 0.686 & 0.624\\
CMT & LC & \textbf{0.729} & \textbf{0.703} \\

\bottomrule
\end{tabular}
\end{center}
\vspace{-1.5em}
\end{table}

\subsection{Decoder and Loss}

As for the decoder, we follow the original transformer decoder in DETR~\cite{wang2021anchor} and use $L$ decoder layers. For each decoder layer, the position-guided queries interact with the multi-modal tokens and update their representations. Two feed-forward networks (FFNs) are used to predict the 3D bounding boxes and the classes using updated queries. We formulate the prediction process of each decoder layer as follows:
\begin{equation}
\hat{b}_i=\Psi^{reg}(Q_{i}), \hat{c}_i=\Psi^{cls}(Q_{i}),
\end{equation}
where $\Psi^{reg}$ and $\Psi^{cls}$ respectively  represent the FFN for regression and classification.
$Q_{i}$ is the the updated object queries of the $i$-th decoder layer.

For set prediction, the bipartite matching is applied for one-to-one assignment between predictions and ground-truths. We adopt the focal loss for classification and $L1$ loss for 3D bounding box regression:
\begin{equation}
L(y,\hat{y})= \omega_{1} L_{cls}(c,\hat{c}) + \omega_{2} L_{reg}(b,\hat{b})
\end{equation}
where $\omega_{1}$ and $\omega_{2}$ are the hyper-parameter to balance the two loss terms. Note that for positive and negative queries in query denoising, the loss is calculated in the same way.

\begin{table*}[tb!]
    \caption{Quantitative results on the nuScenes val with LiDAR or camera miss. With the masked-modal training, the efficiency and robustness of our CMT is significantly improved, especially when the LiDAR camera is missed. 
    }\vspace{-0.2cm}
    \label{table:robust_modal}
    % \resizebox{0.8\textwidth}{!}{
    % \setlength{\tabcolsep}{5pt}
    \centering
    \begin{tabular}{c|ccc|ccc}
        \hline\noalign{\smallskip}
        % \noalign{\smallskip}
        \multirow{2}{*}{Metric} & \multicolumn{3}{c|}{Vanilla training} & \multicolumn{3}{c}{Masked-modal training} \\
        % \cline{2-5}
         & CMT & only LiDAR & only Cams & CMT & only LiDAR & only Cams \\
        % \noalign{\smallskip}
        % \hline\noalign{\smallskip}
        \midrule
        % \noalign{\smallskip}
        NDS $\uparrow$ & 0.726 & 0.603 & 0.073 & 0.729 ($\uparrow$0.3\%) & 0.681 ($\uparrow$\textbf{7.8\%}) & 0.447 ($\uparrow$\textbf{37.4\%})   \\ %($\downarrow$0.042) % ($\downarrow$0.285) 
        mAP $\uparrow$ & 0.691 & 0.487 & 0.000 & 0.703 ($\uparrow$1.2\%) & 0.617 ($\uparrow$\textbf{13.0\%})  & 0.383 ($\uparrow$\textbf{38.3\%})  \\ % ($\downarrow$0.081)  ($\downarrow$0.308) 
        \bottomrule
        \end{tabular}
    % }
\end{table*}

\begin{table}[tb!]
    \caption {NDS/mAP comparison on nuScenes val with sensor miss. BEVFusion is trained with mask-modal strategy. * means our reproduced result. 
    % We use the same image and LiDAR backbones for fair comparison.
    }\vspace{-0.2cm}
    \label{table:robust_comparison}
    % \resizebox{0.8\textwidth}{!}{
    % \setlength{\tabcolsep}{5pt}
    \centering
    \begin{tabular}{c|ccc}
        \hline\noalign{\smallskip}
        % \noalign{\smallskip}
        \multirow{2}{*}{Modal} & \multicolumn{3}{c}{Test modal} \\
        % \cline{2-5}
         & Both & only LiDAR & only Cams \\
        % \noalign{\smallskip}
        % \hline\noalign{\smallskip}
        \midrule
        % \noalign{\smallskip}
        TransFusion~\cite{bai2022transfusion} & 0.71/0.67 & 0.70/0.65 & None  \\
        BEVFusion~\cite{liu2022bevfusion}* & 0.72/0.68 & 0.68/0.63 & 0.40/0.32  \\
        CMT & 0.73/0.70 & 0.68/0.62 & \textbf{0.45/0.38} \\ 
        \bottomrule
        \end{tabular}
    % }
\end{table}

\subsection{Masked-Modal Training for Robustness}
\begin{figure}[h]
	\centering  
\includegraphics[height=2.8cm,width=8cm]{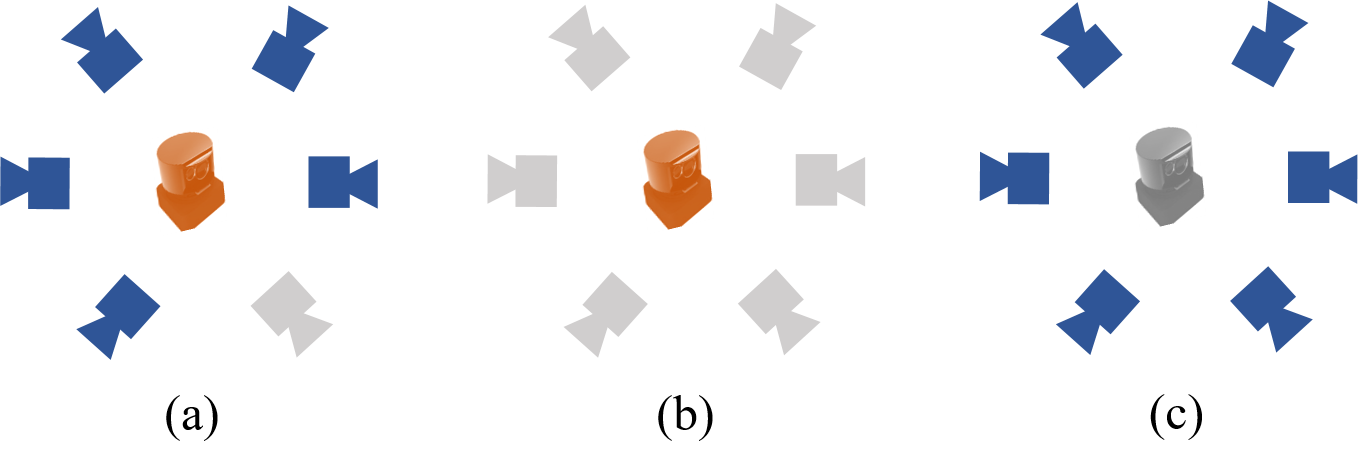}\vspace{-0.2cm}
	\caption{We analyze the system robustness of CMT at test period under three simulated sensor errors: (a) single camera miss, (b) all camera miss and (c) LiDAR miss.
   }  
\label{fig:roboust}
\end{figure}

Security is the most important concern for autonomous driving systems. An ideal system requires solid performance even if part of them fails, as well as not relying on any input of a specific modality.  Recently, BEVFusion~\cite{liang2022bevfusion} has explored the robustness of LiDAR sensor failure. However, the exploration is limited to restricted scan range and model need be retrained. In this paper, we try more extreme failures, including single camera miss, camera miss and LiDAR miss, as shown in \cref{fig:roboust}. It is consistent with the actual scene and ensures the safety of autonomous driving. 

To improve the robustness of the model, we propose a training strategy, called masked-modal training. In training process, we randomly use only a single modality for training, such as camera or LiDAR, with the ratio of $\eta_1$ and $\eta_2$. This strategy ensures that the model are fully trained with both single modal and multi-modal. Then the model can be tested with single modal or multi-modal, without modifying the model weight. The experimental results show that masked-modal training will not affect the performance of our fusion model. Even if LiDAR is damaged, it can still achieve similar performance compared to the SoTA vision-based 3D detectors~\cite{liu2022petr, huang2021bevdet} (see Tab.~\ref{table:robust_modal}-\ref{table:robust_comparison}).

\subsection{Discussion}
CMT shares similar motivation with FUTR3D~\cite{chen2022futr3d} on the end-to-end modeling. However, both the method and its effectiveness are totally different. FUTR3D repeatly samples the corresponding features from each modal and then performs the cross-modal fusion. CMT conducts the position encoding for both multi-view images and point clouds, which are simply added with corresponding modal tokens, removing the repeated projection and sampling processes. It keeps more end-to-end spirits in original DETR framework. Moreover, CMT achieves much better performance compared to the FUTR3D (see comparison in Tab.~\ref{tab:nuscenes_test}), showing its superior effectiveness. We think CMT provides a better end-to-end solution for multi-modal object detection.

\section{Experiments}
%-------------------------------------------------------------------------
\subsection{Datasets and Metrics}
We evaluate our method on open datasets, including nuScenes~\cite{caesar2020nuscenes} and Argoverse 2~\cite{wilson2023argoverse}. 

\textbf{NuScenes}~\cite{caesar2020nuscenes} is a large-scale multi-modal dataset, which is composed of data from 6 cameras, 1 LiDAR and 5 radars. The dataset has 1000 scenes totally and is divided into 700/150/150 scenes as train/validation/test sets, respectively. Each scene has $20s$ video frames with 12 FPS. 3D bounding boxes are annotated every $0.5s$. We only use these key frames. In each frame, nuScenes provides images from six cameras. NuScenes provides a 32-beam LiDAR with 20 FPS. The key frames are also annotated every 0.5s, the same as cameras. We follow the common practice to transform the points from the past 9 frames to the current frame for training and evaluation. We follow the nuScenes official metrics. 

We report the nuScenes Detection Score (NDS), mean Average Precision (mAP), mean Average Translation Error (mATE), mean Average Scale Error (mASE), mean Average Orientation Error(mAOE), mean Average Velocity Error (mAVE) and mean Average Attribute Error (mAAE).

\textbf{Argoverse 2(AV2)}~\cite{wilson2023argoverse} contains 1000 sequences in total, 700/150/150 for train/validation/test similar as nuScenes. AV2 privides a long perceptron range up to $200$ meters, covering an area of $400m \times 400m$, which is much larger than nuScenes. We report mean Average Precision(mAP), Composite Detection Score(CDS).

\begin{table}[tb!]
    \caption {CDS/AP comparison on Argoverse2 \textbf{val} set. ``L" is LiDAR and ``C" is camera.
    % We use the same image and LiDAR backbones for fair comparison.
    }\vspace{-0.2cm}
    \label{table:argoverse2}
    % \resizebox{0.8\textwidth}{!}{
    % \setlength{\tabcolsep}{5pt}
    \centering
    \setlength{\tabcolsep}{3mm}{
    \renewcommand{\arraystretch}{1}
    \begin{tabular}{c|ccc}
        \hline\noalign{\smallskip}
        % \noalign{\smallskip}
        \vspace{1mm}
        Model & Modality & AP & CDS \\
        % \noalign{\smallskip}
        % \hline\noalign{\smallskip}
        \midrule
        % \noalign{\smallskip}
        VoxelNeXt\cite{chen2023voxelnext} & L & 0.307 & -  \\
        FSF\cite{li2023fully} & LC & 0.332 & 0.255  \\
        CMT & LC & \textbf{0.361} & \textbf{0.278} \\ 
        \bottomrule
        \end{tabular}}
    % }
\end{table}

\subsection{Implementation Details}
\label{implementation}
We use ResNet\cite{he2016resnet} or VoVNet\cite{lee2020centermask} as image backbone to extract the 2D image features.
The C5 feature is upsampled and fused with C4 feature to produce P4 feature. We use VoxelNet~\cite{zhou2018voxelnet} or PointPillars~\cite{lang2019pointpillars} as the backbone to extract the point-cloud features. 
We set the region-of-interest (RoI) to $[-54.0m, 54.0m]$ for X and Y axis, and $[-5.0m, 3.0m]$ for Z axis. The 3D coordinates in the world space are normalized to $[0, 1]$. All the feature dimension is set to 256, including the LiDAR feature, image feature and query embedding. Six decoder layers are adopted in transformer decoder. Voxel size of 0.075 and image size of $1600\times640$ are adopted as default in our experiments.

Our model is trained with the batch size of 16 on 8 A100 GPUs. It is trained for total 20 epochs with CBGS\cite{zhu2019class}.
We adopt the AdamW\cite{loshchilov2017decoupled} optimizer for optimization. The initial learning rate is $1.0\times 10^{-4}$ and we follow the cycle learning rate policy\cite{smith2017cyclical}. The mask ratios $\eta_1$ and $\eta_2$ are both set to 0.25 for masked-modal training. 
% The threshold $\xi$ is set to 0.75 to divide the noise queries into positives and negatives for training. The tolerance $\lambda$ that controls the noise scale is set to 1. 
The GT sample augmentation is employed for the first 15 epochs and closed for the rest epochs. As for the loss weights, we follow the default setting in DETR3D~\cite{wang2022detr3d} and set the $\omega_{1}$ and $\omega_{2}$ to 2.0 and 0.25, respectively.
For fast convergence, we introduce the point-based query denoising strategy based on DN-DETR~\cite{li2022dn}. Different from it, we generate the noisy anchor points by center shifting since the box scale is not that important in 3D object detection.

\begin{figure*}[t]
\centering  
\includegraphics[width=0.95\textwidth]{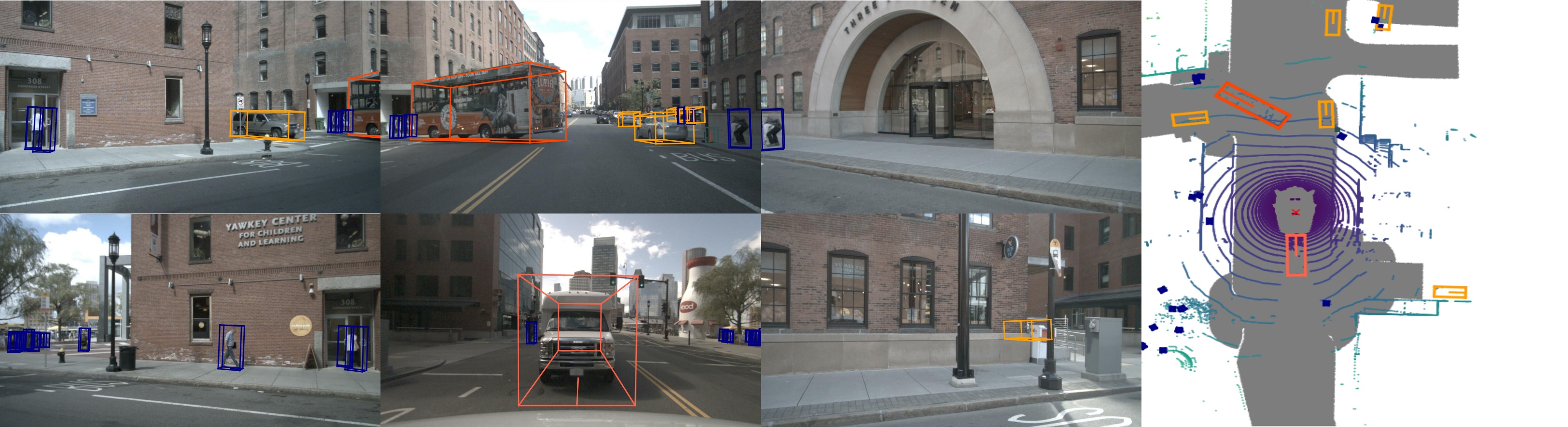}\vspace{-0.2cm}
\caption{Some qualitative detection results on the surrounding views and BEV space in the nuScenes test set. Bounding boxes with different colors represent vehicles($\mycirc[car_color]$), pedestrians($\mycirc[pedestrian_color]$), Bus($\mycirc[bus_color]$) and Truck($\mycirc[truck_color]$). 
% CMT can precisely detect the surrounding objects.
} 
\label{fig:visualize_case}
\end{figure*}

On AV2, our model is trained 6 epochs, following common practice\cite{chen2023voxelnext, li2023fully}. 

\begin{figure}[t]
\centering  
\includegraphics[width=0.45\textwidth]{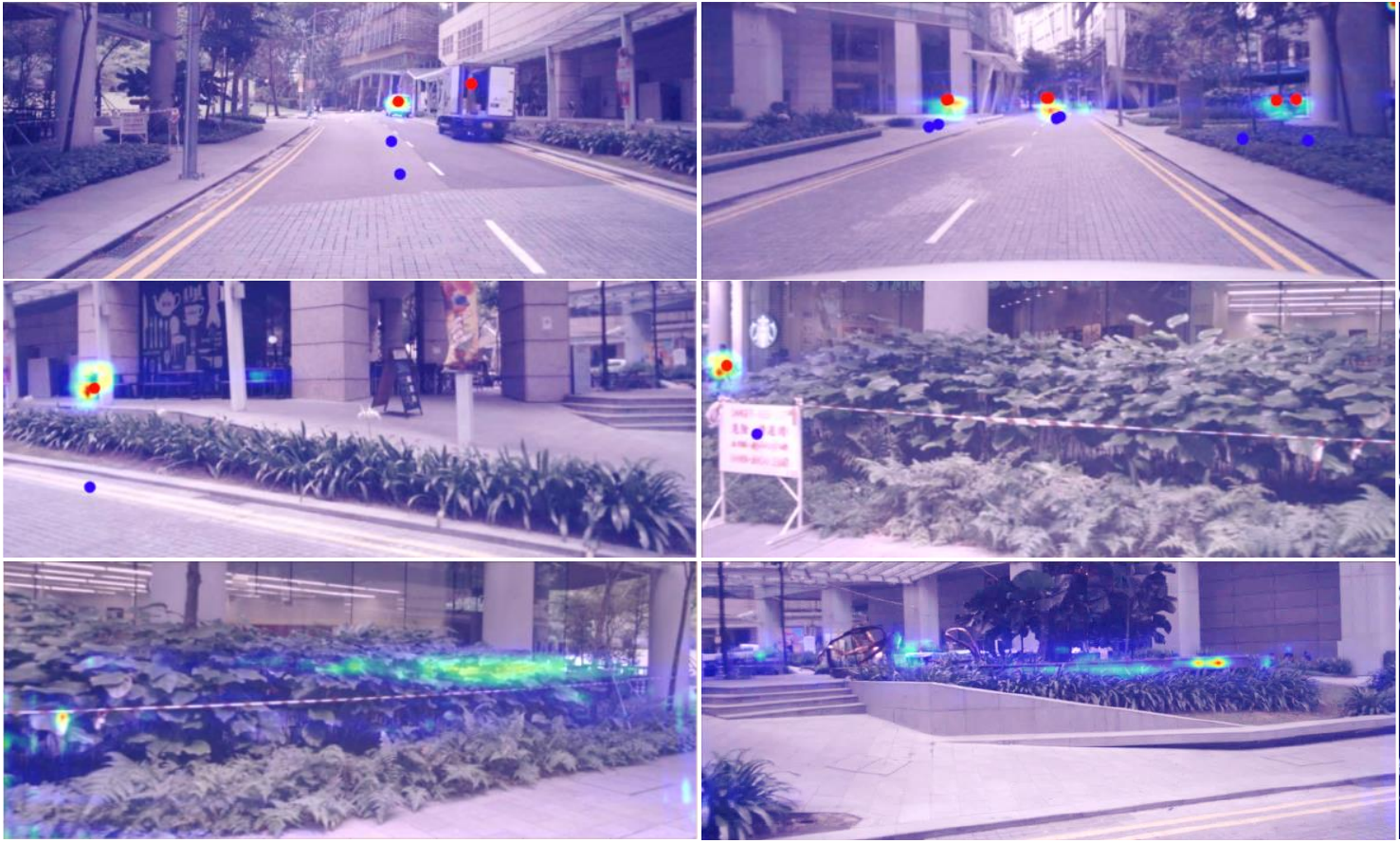}
\vspace{-0.2cm}
\caption{Visualization of attention maps on multi-view images. The blue points ($\mycirc[blue]$) are initial anchor points while red points ($\mycirc[red]$) are the centers of box predictions. It shows that high response regions of attention maps mainly focus on the foreground objects, which are close to the anchor points.}  
\label{fig:heatmap}
\end{figure}

\subsection{ State-of-the-Art Comparison}
As shown in \cref{tab:nuscenes_test}, CMT achieves state-of-the-art performance compared to existing methods on nuScenes test set. Our LiDAR-only baseline, named CMT-L, achieves 70.1\% NDS, which is a nearly SoTA performance among all existing LiDAR-only methods.
Our multi-modal CMT achieves 74.1\% NDS and 72.0\% mAP, outperforming all existing SoTA approaches, such as BEVFusion~\cite{liu2022bevfusion} and DeepInteration~\cite{yang2022deepinteraction}. 
% Benefiting from the large receptive field, CMT gains better results on some metrics like mAVE.
We also compare the performance with other SoTA methods on nuScenes val set (see \cref{tab:nuscenes_val}). It shows that our proposed CMT with multi-modal fusion outperforms the BEVFusion by 1.8\% mAP. CMT introduces large performance improvements compared to our LiDAR-only CMT-L by 4.0\%/6.7\% and 4.3\%/7.9\% NDS/mAP on test and validation set, respectively. In comparison, TransFusion only brings 1.5\%/3.4\% NDS/mAP on test set, compared to the LiDAR-only TransFusion. It shows that the multi-view images bring much more complementary information to the point clouds in CMT framework. We think the end-to-end modeling of CMT relatively improves the importance of image tokens. Fig.~\ref{fig:visualize_case} shows some qualitative detection results on the nuScenes test set.

\begin{table*}[thb]
    % \centering
    \begin{center}
    \caption{
    The ablation studies of different components in the proposed CMT. 
    }
    \vspace{-0.2cm}
    \label{tab:ablation}
    \quad\quad\ 
    \setlength{\tabcolsep}{2pt}
    \begin{subtable}[t]{0.45\linewidth}
        \label{tab:ablation_query}
        \begin{tabular}{ll|ccccc}
        \hline\noalign{\smallskip}
         \thinspace\enspace Im \quad & \quad PC\quad & NDS & mAP & mATE & mASE & mAOE\\
        \noalign{\smallskip}
        \hline
        \noalign{\smallskip}
        \enspace\checkmark& & 0.595 & 0.554 & 0.515 & 0.258 & 0.429\\
        \enspace&\quad\checkmark & 0.665 & 0.626 & \textbf{0.372} & \textbf{0.255} & \textbf{0.347}\\
        \enspace\checkmark&\quad\checkmark & \textbf{0.669} & \textbf{0.641} & 0.377 & 0.254 & 0.375\\ 
        \bottomrule
        \end{tabular}
        \caption{Position encoding for query.}
    \end{subtable}
    % \hspace{2mm}
    \setlength{\tabcolsep}{2pt}
    \begin{subtable}[t]{0.45\linewidth}
        \label{tab:ablation_denoise}
        \begin{tabular}{l|ccccc}
        \hline\noalign{\smallskip}
         \quad\enspace\thinspace PQD\quad\hspace{1mm}& NDS & mAP & mATE & mASE & mAOE \\
        \noalign{\smallskip}
        \hline
        \noalign{\smallskip}
         & 0.626 & 0.584 & 0.429 & 0.259 & 0.420\\
        \quad\quad\enspace\checkmark\quad & \textbf{0.669} & \textbf{0.641} & \textbf{0.377} & \textbf{0.254} & \textbf{0.375} \\ 
        % \hline
        \bottomrule
        \end{tabular}
        \caption{Point-based query denoising.}
    \end{subtable}
    % \hspace{5mm}
    
    \quad\quad\ 
    \setlength{\tabcolsep}{2.0pt}
    \begin{subtable}[t]{0.45\linewidth}
        \label{tab:abalation_voxelsize}
        \begin{tabular}{l|ccccc}
        \hline\noalign{\smallskip}
        % \qquad\qquad\qquad & NDS & mAP & mATE & mASE & mAOE \\
        Voxel size & NDS & mAP & mATE & mASE & mAOE \\
        \noalign{\smallskip}
        \hline
        \noalign{\smallskip}
        \quad0.075\quad\enspace & 0.669 & \textbf{0.641} & \textbf{0.377} & 0.254 & 0.375\\
        \quad0.1\quad\enspace & \textbf{0.671} & 0.638 & 0.378 & \textbf{0.252} & \textbf{0.334}\\
        \quad0.125\quad\enspace & 0.655 & 0.624 & 0.396 & 0.255 & 0.397 \\
        \bottomrule
        \end{tabular}
        \caption{Voxel size of LiDAR backbone.}
    \end{subtable}
    % \hspace{5mm}
    \setlength{\tabcolsep}{2pt} 
    \begin{subtable}[t]{0.45\linewidth}
        \label{tab:abalation_imgbackbone}
        \begin{tabular}{l|ccccc}
        \hline\noalign{\smallskip}
        % \quad \quad & NDS & mAP & mATE & mASE & mAOE \\ 
        Backbone & NDS & mAP & mATE & mASE & mAOE \\ 
        \noalign{\smallskip}
        \hline
        \noalign{\smallskip}
        ResNet-50 & 0.658 & 0.623 & \textbf{0.376} & \textbf{0.253} & 0.399 \\
        ResNet-101 & 0.664 & 0.629 & 0.383 & 0.254 & \textbf{0.363} \\
        VoV-99 & \textbf{0.669} & \textbf{0.641} & 0.377 & 0.254 & 0.375\\
        \bottomrule
        \end{tabular}
        \caption{Image backbone.}
    \end{subtable}
    % \hspace{3mm}
    
    \quad\quad\ 
    \hspace{-1mm}
    \thinspace
    \setlength{\tabcolsep}{2pt}
    \begin{subtable}[t]{0.45\linewidth}
        \label{tab:abalation_imgsize}
        \begin{tabular}{l|ccccc}
        \hline\noalign{\smallskip}
        % \quad \quad & NDS & mAP & mATE & mASE & mAOE \\
        Image size & NDS & mAP & mATE & mASE & mAOE \\
        \noalign{\smallskip}
        \hline
        \noalign{\smallskip}
        \thinspace $800\times 320$ & 0.654 & 0.609 & \textbf{0.374} & 0.256 & 0.389\\
         $1600\times 640$ & \textbf{0.669} & \textbf{0.641} & 0.377 & \textbf{0.254} & \textbf{0.375}\\
        \bottomrule
        \end{tabular}
        \caption{Input size of image backbone.}
    \end{subtable}
    % \hspace{5mm}
    \thinspace
    \setlength{\tabcolsep}{2pt} 
    \begin{subtable}[t]{0.45\linewidth}
        \label{tab:abalation_lidarbackbone}
        \begin{tabular}{l|ccccc}
        \hline\noalign{\smallskip}
        % \quad \quad & NDS & mAP & mATE & mASE & mAOE \\ 
        Backbone & NDS & mAP & mATE & mASE & mAOE \\ 
        \noalign{\smallskip}
        \hline
        \noalign{\smallskip}
        PointPillars\thinspace & 0.628 & 0.598 & 0.430 & \textbf{0.252} & 0.455 \\
        VoxelNet\thinspace & \textbf{0.669} & \textbf{0.641} & \textbf{0.377} & 0.254 & \textbf{0.375} \\
        \bottomrule
        \end{tabular}
        \caption{Lidar backbone}
    \end{subtable}
    \label{tab:ablation}
\end{center}
\vspace{-2.0em}
\end{table*}

\subsection{Strong Robustness}
We evaluate the robustness of our framework under various harsh environments, including LiDAR miss and camera miss. 
\cref{table:robust_modal} shows the results when the sensor miss occurs, by simulating the scenarios of any modality totally broken. 
The performance is compared between the vanilla training and masked-modal training. 
It validates the effect of masked-modal training. Note that the model are only trained with multi-modality and evaluated without any finetune process.
With vanilla training, the model fails to predict anything meaningful (only Cams with mAP=0) when LiDAR is missing.
With masked-modal training, the absence of LiDAR or camera modalities lead to $4.8$\% and $28.2$\% NDS drop compared to CMT, respectively. It is observed that losing one modality still remains similar results compared to single-modal training settings. It overcomes the drawback that multi-modal method usually rely on one major modality and performance would degrade significantly if losing the major modality. Especially, for the case of LiDAR missing, the performance is still comparable to the SoTA camera-only method PETR~\cite{liu2022petr}, validating the strong robustness of our method. 
We further evaluate the performance of TransFusion and BEVFusion under sensor miss (see Tab.~\ref{table:robust_comparison}). TransFusion fails to work when LiDAR is missing due to the two-stage design. With the masked-modal training, BEVFusion achieves the decent performance (40\% NDS and 32\% mAP), while showing large inferiority compared to CMT. 

Moreover, we also investigate the case when any one of cameras fails. Experimental result shows slight performance drop, indicating the tolerable to single camera miss of our method. Six sensors brings an average decrease of 0.7\% NDS, no more than 1\% performance of the oracle version. The front and back sensor relatively play the important role among camera sensors, with 1.1\% and 0.8\% decrease respectively, due to their distant or large field of view. Compared to the camera-only setting, our multi-modal framework facilitate the compensation between LiDAR and image domains, thus presenting a robust performance. 

\subsection{Ablation Study} 
We present the ablation studies in \cref{tab:ablation}. All the experiments are conducted for 20 epochs without CBGS\cite{peng2018megdet}. 
We first ablate the effect of Im PE and PC PE on the generation of position-guided queries. 
It shows that removing PC PE introduces a $7.4\%/8.70\%$ NDS/mAP performance drop, which is much larger than the drop of removing Im PE $0.4\%/1.5\%$.
Next, we explore the effectiveness of point-based query denoising (PQD) introduced in Sec.~\ref{implementation}. We can easily find that PQD can greatly improve the overall performance by $4.3\%/5.7\%$ NDS/mAP. With PQD, the  training convergence can be boosted, which is similar to the practice in DN-DETR~\cite{li2022dn}.
Further, we also illustrate the effect of scaling up the CMT model as well as the input size. Overall, CMT can benefit from the scaling model size. Interestingly, we find increasing the voxel number (smaller voxel size) and image size achieves similar improvements $\approx1.5\%$ in NDS. While scaling the image size increases more mAP than the voxel number(+3.2\% \vs +1.7\%). When increasing the image size from $800\times 320$ to $1600 \times 640$, we find the performance improvements are mainly from these small objects, such as pedestrian and motorcycle. 
We also conduct experiments on replacing image and LiDAR backbones, we use VoV-99\cite{lee2020centermask} and ResNet\cite{he2016resnet} as our image backbones. 
Experiments show that our proposed CMT can benefit from larger backbones. For image, VoV-99 backbone achieves the best result and outperforms the ResNet-50 by $1.1\%/1.8\%$ in NDS/mAP. While for LiDAR, VoxelNet outperforms the PointPillar by $4.1\%/4.3\%$ in NDS/mAP.

\subsection{Analysis}
% CMT is a direct and easy pipeline for multi-modal fusion and can be easily extended. Moreover, benefiting from DETR\cite{carion2020detr} framework and our training schedule, CMT shows strong robustness under sensor miss conditions. We present some attempted experiments in this section.

% \noindent \textbf{Data extension.} 
% Multi-frame is now a common setting in camera-based 3D object detection\cite{liu2022petrv2, huang2022bevdet4d, li2022bevformer}. Using multi frames often outperforms the single frame by a clear margin and can solve some typical occlusion problem. We follow the multi-frame alignment in PETRv2\cite{liu2022petrv2}. Considering the high memory cost of multi-frames, we conduct our experiment with a $800\times 320$ image resolution. As shown in \cref{tab:abalation_frame_radar}, adding image frame only improves the NDS/mAP by $0.2\% / 0.7\%$. More image frames rarely improve the performance with multi-modal fusion. 

% Radar has advantages in long range detection and the robustness of extreme weather. Following FUTR3D\cite{chen2022futr3d}, we stack points from 5 radars together to generate the point cloud. We use several MLP layers to perform coordinates encoding on Radar features, the same as LiDAR. \cref{tab:abalation_frame_radar} shows that adding Radar data to our pipeline degrades the performance by 0.9\% NDS and 0.6\% mAP.

For better understanding on querying from multi-modal tokens, we visualize the attention map of cross-attention on the multi-view images (see \cref{fig:heatmap}). We can clearly find that the attention maps have higher response on the regions that includes foreground objects. It shows that our method can implicitly achieve the cross-modal interaction. We visualize the initial anchor points and the center points of predictions. Most anchor points focus on the closest foreground objects. After the interaction with multi-modal tokens in the transformer decoder, anchor points are updated and gradually access the accurate center points.

% \begin{table}[h]
% \begin{center}
% \caption{Results with more image frames or with Radar points. 
% % The metrics is NDS/mAP.
% }\vspace{-0.2cm}
% \label{tab:abalation_frame_radar}
% \begin{tabular}{cc|cc}
% \hline\noalign{\smallskip}
% +Frame &  + Radar & \ NDS\ & \ mAP\ \\
% \midrule
% & & \ 0.698\ & \ 0.662\ \\
% \checkmark & & \ \textbf{0.700}\ & \ \textbf{0.669}\ \\
%  & \checkmark & \ 0.689\ & \ 0.656\ \\
% \bottomrule
% \end{tabular}
% \end{center}
% \vspace{-1.5em}
% \end{table}

\section{Conclusions}
In this paper, we propose a fully end-to-end framework for multi-modal 3D object detection. It implicitly encodes the 3D coordinates into the tokens of images and point clouds. With the coordinates encoding, the simple yet effective DETR pipeline can be adopted for multi-modal fusion and end-to-end learning. With masked-modal training, our multi-modal detector can be learned with strong robustness, even if one of multi-modalities are missed. We hope such a simple pipeline design could provide more insights on the end-to-end 3D object detection.

\noindent \textbf{Acknowledgements:} This research was supported by National Key R\&D Program of China (No. 2017YFA0700800) and Beijing Academy of Artificial Intelligence (BAAI).

% \noindent \textbf{Limitation:} Though our CMT brings some advantages compared to those existing approaches, it also reveals some limitations. The computation cost is relatively larger due to the large number of multi-modal tokens and the global attention employed in transformer decoder. 
% To solve this problem, some  efforts in two directions maybe taken. The first one is to reduce the redundancy of multi-modal tokens. The foreground tokens can be roughly selected with another individual network~\cite{wang2021adaptive}. The foreground tokens are then input to our network for high-speed inference. Another possible solution is to replace the global attention with other efficient attentions, like deformable attention~\cite{zhu2020deformable}. 
% One can also employ a small set of object queries since most queries correspond to the empty objects. Recently, FlashAttention~\cite{dao2022flashattention} figures out the speed of most Transformers are restricted by memory access. Naive attention, employed in our CMT framework, can greatly benefit from making fully use of fast on-chip SRAM.
% carefully accounting for reads and writes between fast on-chip SRAM and GPU HBM, 

{\small
\bibliographystyle{ieee_fullname}
\bibliography{egbib}
}

\end{document}